\documentclass[runningheads]{llncs}
\usepackage[T1]{fontenc}
\usepackage{graphicx}
\usepackage[dvipsnames]{xcolor}
\usepackage[ruled,vlined]
{algorithm2e}
\usepackage{subcaption}
\usepackage{booktabs}
\usepackage{amsmath}
\usepackage{url}

\usepackage[normalem]{ulem}

\begin{document}
\raggedbottom
\title{Future Querying: Can LLMs Serve as Implicit Medical World Models?}

\author{
Siri Willems\textsuperscript{*} \and
James Butterworth\textsuperscript{*}\and
Lore Goetschalckx\textsuperscript{*} \and
Peter Vrancx\textsuperscript{*} \and
Philippe Modard \and
Elke Giets \and
Ludovic Denoyer
}
\authorrunning{S. Willems et al.}
\institute{imec, AI-labs\footnote{\url{http://ailabs.imec-int.com}}, Paris, France}
\maketitle
\let\thefootnote\relax\footnotetext{\textsuperscript{*}Equal contribution}
\begin{abstract}
%Traditional clinical prediction models rely on task-specific pipelines and curated, structured data, which scale poorly and underutilize unstructured text. To address this, we introduce future querying, a paradigm that tests whether large language models (LLMs) can function as implicit medical world models. Our framework operates directly on unstructured clinical documentation using task-agnostic training, enabling flexible inference over patient trajectories without manual feature engineering. Evaluated on a new synthetic medical reports datasets and real ICU notes from the MIMIC-IV dataset, our framework demonstrates strong potential as a general-purpose method and moreover is suitable for on-premise deployment.
Traditional clinical prediction models rely on task-specific pipelines and curated, structured data, which scale poorly and underutilize unstructured text. To address this, we introduce future querying, a paradigm that probes whether large language models (LLMs) can function as implicit medical world models by evaluating their ability to answer time-indexed clinical queries about a patient's future. Our framework operates on unstructured clinical documentation using endpoint-agnostic training, enabling a single model to answer diverse clinical queries over patient trajectories without manual feature engineering or task-specific retraining. We show that small, locally fine-tuned open-weight models can match or approach larger proprietary systems, making the framework suitable for privacy-preserving, on-premise deployment. Evaluated on a new synthetic medical reports dataset and real ICU notes from the MIMIC-IV dataset, our results provide encouraging evidence that LLMs can capture aspects of clinical dynamics.% though distinguishing genuine world modeling from pattern matching remains an open question.

\end{abstract}

\section{Introduction}

Clinical decision-making can greatly benefit from an accurate prediction of the future evolution of a patient’s condition.
% In machine learning, this is commonly formulated as predicting future outcomes conditioned on historical data.
Existing machine learning approaches for predicting patient outcomes focus on predefined target outcomes at fixed horizons and rely on task-specific pipelines built on curated structured inputs, e.g., for predicting in-hospital mortality, length of stay, or readmission risk \cite{caruana2015intelligible,harutyunyan2019multitask,Sabouri2023}.
Such approaches exhibit important limitations  
\cite{Kraljevic2024_Foresight,Pellegrini2025_EHR2Path,rajkomar2018scalable,alaa2019attentive,yang2023transformehr} and require training separate models for each target outcome, leading to high development and maintenance costs.
They also depend on restricted sets of manually selected variables, potentially discarding informative signals and limiting their ability to capture complex patient trajectories.
Moreover, large amounts of clinically relevant information in unstructured data (e.g., medical reports) remain underutilized or require costly extraction pipelines. 

We investigate whether large language models (LLMs) can instead serve as implicit medical world models through a paradigm we term future querying. In this paradigm, forecasting is formulated as answering flexible, time-indexed natural language queries about a patient’s future, conditioned on the patient’s observed trajectory. A model capable of answering such queries must capture the dynamics governing the evolution of patient state over time, providing evidence that it may implicitly encode a medical world model.
We focus on textual models operating on unstructured clinical documentation (e.g., longitudinal clinical notes and medical reports), which provide a unified, temporally ordered representation of patient state while implicitly capturing information derived from multiple clinical modalities.

Our approach departs from prior work in two key aspects.
First, it requires no feature engineering or structured preprocessing, operating directly on raw clinical text.
Second, the approach is task-agnostic, allowing a single model to answer diverse clinical queries without task-specific retraining.
Having a task-agnostic approach  potentially unlocks multiple downstream use-cases with a single model, such as automatic diagnosis, counter-factual treatment simulations and mortality prediction.
By fine-tuning pre-trained LLMs locally, we obtain a unified and privacy-preserving framework suitable for on-premise deployment.
We evaluate our approach on synthetic medical reports we generate as well as real intensive care unit (ICU) notes from the MIMIC-IV dataset \cite{PhysioNet-mimiciv-2.2,Johnson2023}, demonstrating its potential as a general-purpose method for modeling patient trajectories.

\section{Related work}
The question of whether LLMs learn implicit world models is an active topic of debate in the AI community. Several studies provide evidence that large-scale training for next-token prediction induces learning of internal world models \cite{li2021implicit,li2023othello,jin2024emergent,levy2025worldllm}. Other authors note that predictive accuracy does not necessarily imply the presence of a consistent world model and that LLM representations can often be incoherent \cite{vafa2024evaluating,li2024llm_world_representations}. In more empirical settings, \cite{halawi2024forecasting,chuang2025woc} demonstrated that LLMs are capable of real-world future forecasting, though these results are very methodology dependent, as \cite{schoenegger2023forecasting} find that LLM predictions perform no better than random chance in their setup.

In the medical domain, prior work has explored sequence models as predictors of patient health trajectories. These works focus on training task-specific medical forecasting models, rather than leveraging medical world models induced in pre-trained LLMs. Early research focuses on forecasting patient outcomes from longitudinal electronic health records (EHRs), which record sequences of clinical events over time \cite{allam2021analyzing}. These approaches operate on structured data and target predefined endpoints such as mortality, readmission, or diagnosis prediction, using deep learning models including transformer-based architectures \cite{yang2023transformehr,rong2025deep,Makarov2025,Kraljevic2024_Foresight,Pellegrini2025_EHR2Path}. In parallel, several studies have shown that unstructured clinical notes can support prognostic modeling in ICU settings and beyond by extracting structured context from unstructured notes, combined with classical machine learning approaches \cite{mahbub2022unstructured,zaghir2021real}. 

\section{Task formulation}
We describe a patient $p$ at current time $T$ by a timeline of medical reports
\begin{equation}
p = \{(t_1, r_1), \ldots, (t_n, r_n)\}
\end{equation}

where each timestamp $t_i \leq T$ corresponds to a report $r_i$, represented as unstructured text that can potentially be extracted from PDFs or scanned documents. A classical world modeling approach would aim to predict future reports for times $t > T$. However, such an objective is inherently ill-posed, as it requires the model not only to capture the patient's underlying clinical evolution, but also to reproduce irrelevant artifacts such as writing styles or reporting conventions.

To address this limitation, we consider a simpler and more practical formulation: we posit that a model $f$ that can accurately answer questions about a patient’s future given their past trajectory has likely captured aspects of the underlying clinical dynamics, a necessary, though not sufficient, condition for acting as a medical world model. Under this perspective, the goal is no longer to generate future reports, but to enable informed querying of future outcomes. We consider a \emph{query} $q \in \mathcal{Q}$ that is issued at time $T$ and asks about the state of the patient $p$, expressed in natural language and associated with a %a target time $t$. A query can be seen as a combination of (i) a time-agnostic textual predicate $x$ describing \emph{what} is being asked, and (ii) a time horizon $\Delta t = T-t$ specifying \emph{when} the information is sought relative to the present.
target time $t \ge T$. A query can be seen as a combination of (i) a time-agnostic textual predicate $x$ describing \emph{what} is being asked, and (ii) a timestamp $t$ specifying \emph{when} the information is sought. 

Our objective is to build a model $P(a|p,q)$ that models the distribution of answers $a$ given a patient history and a query. Such a model would allow clinicians to pose questions such as \emph{``What is the risk of complication $X$ within in three days?''}, \emph{``Will the patient recover within the next week?''}, or \emph{``Is the patient likely to be discharged soon?''}. The answers to such queries are clinically relevant and can inform both care planning and resource allocation.

\paragraph{Model fine-tuning.}

Beyond evaluating the medical world knowledge of frontier models, we also examine whether smaller open-weight models can be fine-tuned within this querying framework to improve their medical forecasting ability.

Let us consider a model $f$ with parameters $\theta$ that approximates the distribution $P(a \mid p, q)$ defined above through $f(a \mid p, q; \theta)$. Since $a$, $p$, and $q$ are all expressed as text, it is natural to leverage LLM architectures to capture this distribution. To this end, we adopt a task-agnostic training strategy in which a dataset of patients is transformed into a dataset of training examples $(a, p, q)$, enabling standard supervised fine-tuning methods for the LLM. Assuming access to such a dataset $\mathcal{D} = \{(a_k, p_k, q_k)\}_{k=1}^N$, the model can be trained using a standard maximum likelihood objective:
\begin{equation}
\theta^* = \arg\max_{\theta} \frac{1}{N} \sum_{k=1}^N \log f(a_k \mid p_k, q_k; \theta)
\end{equation}

This optimization can be carried out using standard LLM training procedures, making the approach straightforward to implement and practical to deploy in clinical environments where complex training pipelines may not be feasible.

\section{Building datasets for future querying}\label{sec:building_datasets} 

\begin{figure}[t]
    \centering
    \includegraphics[width=0.70\linewidth]{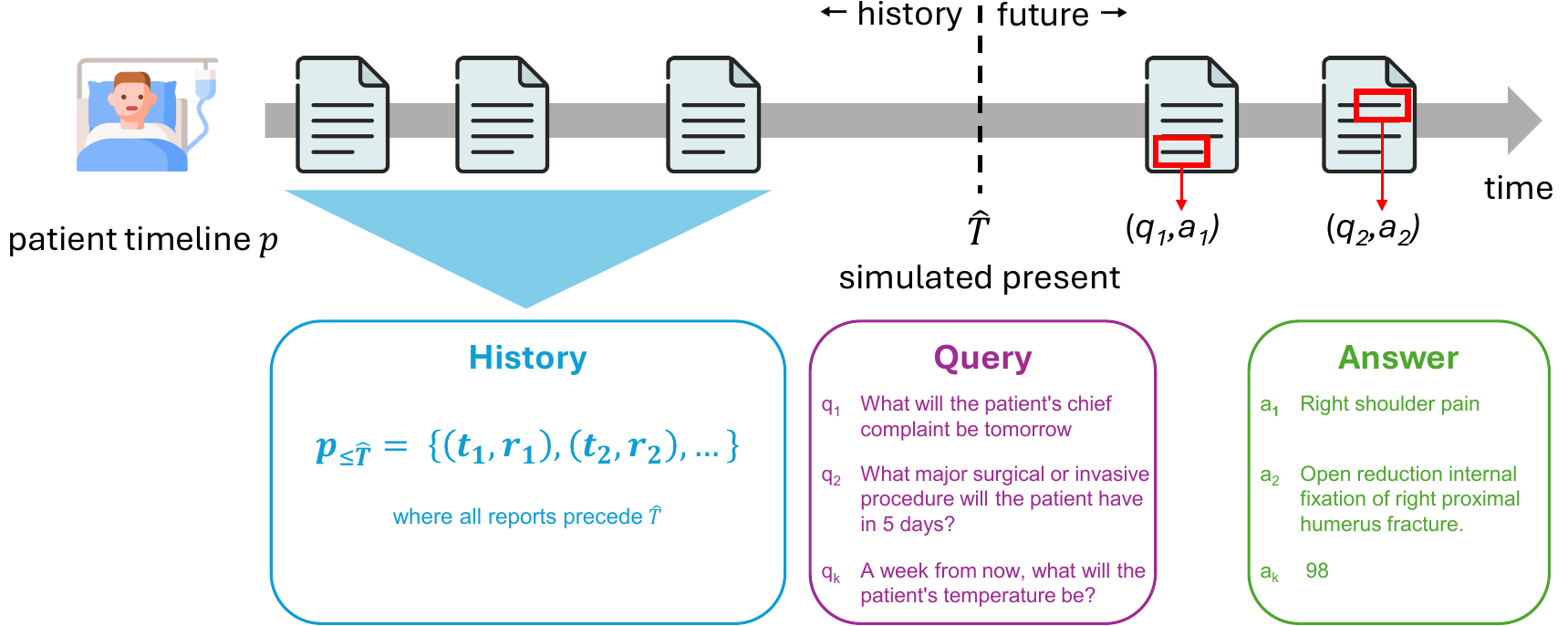}
    \caption{The future querying paradigm. A patient timeline $p$ is partitioned by a virtual present $\hat{T}$ into an observed history and a hidden future. Given the historical context and a flexible, time-indexed clinical query $q_k$, we investigate whether an LLM can serve as an implicit medical world model to predict the correct answer $a_k$ derived from the patient's actual future timeline.}
    \label{fig:queries}
\end{figure}

Clinical information systems rarely provide native, structured datasets for question-answering tasks based on patient histories. Consequently, we develop a method to construct training and validation sets from raw observational data, illustrated in Fig.~\ref{fig:queries}. Let us assume access to a comprehensive corpus of longitudinal patient histories, denoted as: $
\mathcal{P} = \left\{ p_j = \left[ (t_1^j, r_1^j), \ldots, (t_{n_j}^j, r_{n_j}^j) \right] \right\}_{j \in [1, M]}
$. Our objective is to systematically derive the target dataset $\mathcal{D}$ from $\mathcal{P}$. To this end, we use LLMs to parse and synthesize clinical narratives.  Our dataset construction algorithm randomly samples patients $p_j$ from the population $\mathcal{P}$ and proceeds in three key steps for each sampled patient:

\paragraph{Predicate extraction.} Given a random report $r_i^j$ from the patient's timeline, the algorithm formulates predicate questions $x$ about the events or findings discussed within it, along with a corresponding answer $a$ for each. 

\paragraph{Temporal partitioning.} For each predicate, it then selects a random pivot time\-stamp $\hat{T}$ that precedes report time $t_i^j$ within the patient's timeline ($t_1^j \leq \hat{T} < t_{i}^j$). This pivot acts as the simulated ``present'', positioning $t_i^j$ forward in time and casting the final query $q(x, t)$ as one about the patient's future, while also defining its historical context (all reports before $\hat{T}$).

\paragraph{History-query-answer generation.} Once a natural language query $q$ is  formulated based on $x$ and $t$, the resulting tuple, consisting of the historical context, the query $q$, and the answer $a$, undergoes a final validation. Those tuples whose answer is obvious from the historical context alone are not genuinely future querying and are rejected. All accepted tuples are appended to our final future querying dataset $\mathcal{D}$.

\section{Experiments}
\subsection{Evaluation datasets}

We conduct our evaluation on datasets constructed from two distinct patient corpora ($\mathcal{P}$): (i) a synthetic corpus of patient histories that we generate, spanning a  wide range of common medical conditions, and (ii) real-world ICU clinical notes extracted from the publicly available MIMIC-IV dataset \cite{PhysioNet-mimiciv-2.2,Johnson2023}. 

\paragraph{Synthetic Medical Reports.}
We autoregressively generate synthetic patient timelines %histories
using a general-purpose LLM (Gemini 2.5 Flash \cite{gemini25}). %To generate realistic patient timelines%histories,
We first sample a clinically grounded patient profile (disease, demographics, comorbidities) that acts as a latent conditioning variable. Guided by this hidden profile and the chronological sequence of prior reports, the LLM iteratively predicts the next medical report. At each step, the model reasons about visit timing, specialties, examinations, and treatments. Generation terminates when the patient is cured, deceased, or a predefined timeline bound is reached.

\paragraph{MIMIC ICU Notes.}
We complement the synthetic dataset with de-identified intensive care records from MIMIC-IV \cite{PhysioNet-mimiciv-2.2,Johnson2023}. For each patient, we extract time-stamped clinical notes (discharge summaries and radiology reports), sort them chronologically, and convert them into longitudinal patient timelines %histories
annotated with timestamps.
%, grounding our modeling in the authentic temporal structure of real clinical practice.

%\hfill

\begin{figure*}[t]
 \centering
 \begin{minipage}{0.48\columnwidth}
 \centering
 \resizebox{\columnwidth}{!}{%
 \small
 \begin{tabular}{lcc}
 \toprule
  & \textbf{Synthetic Medical} & \textbf{MIMIC ICU} \\
  & \textbf{Reports} & \textbf{Notes} \\
 \midrule
 Patients & 3{,}226 & 9{,}733 \\
 Queries & 53{,}612 & 164{,}911 \\
 Queries / patient & 16 (3--39) & 19 (1--20) \\
 Query history (reports) & 2 (1--9) & 3 (1--237) \\
 Total trajectory (reports) & 10 (2--10) & 14 (2--341) \\
 Report length (chars) & 3{,}958 (53--9{,}440) & 902 (12--54{,}809) \\
 \bottomrule
 \end{tabular}%
 }
 \end{minipage}%
 \hfill
 \begin{minipage}{0.48\columnwidth}
 \centering
 \includegraphics[width=\columnwidth]{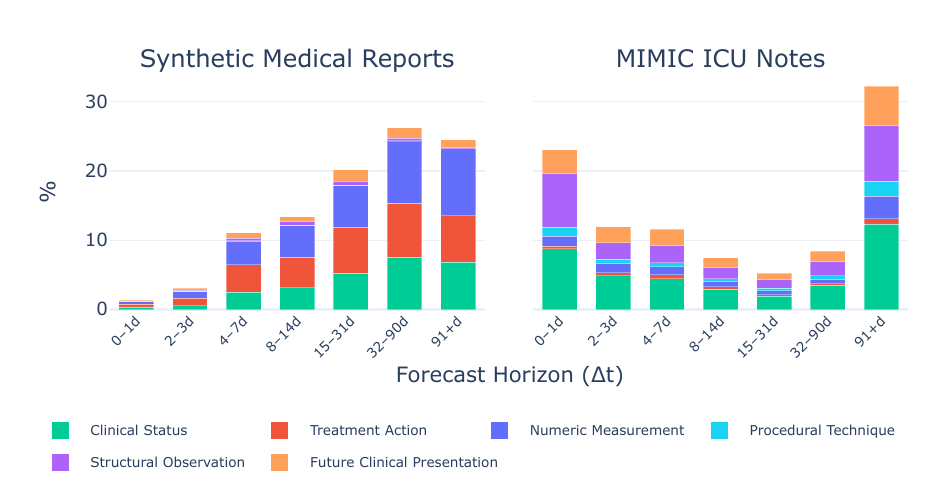}
 \end{minipage}
 \caption{(Left) Characteristics of the two future querying %evaluation
 datasets; values are reported as median (min--max). (Right) Distribution of query types computed on the test fold. Queries span six semantic categories and are binned by the time gap between the simulated present and the target clinical event.}
 \label{fig:dataset-overview}
\end{figure*}

Using the algorithm explained in Section~\ref{sec:building_datasets}, with Gemini 2.5 Flash \cite{gemini25} chosen as the LLM, %to perform predicate and answer extraction, query formulation, and validation, 
we produce tuples of patient history, query, and answer from these two patient corpora. %The prompts used for dataset construction are provided in Appendix~X. 
The outcome is two datasets, which are described in Fig.~\ref{fig:dataset-overview} (left) and (right), each randomly split (patient-wise) into train, validation, and test folds using an 80-10-10 \% ratio.

\subsection{Baselines and models}

To assess the extent to which LLMs can act as implicit medical world models, we evaluate their baseline performance as well as the impact of finetuning LLMs within the future querying framework. Concretely, we measure their ability to answer queries about a patient's future state when they are (i) taken off-the-shelf (i.e., pre-trained), and (ii) fine-tuned with supervision on $(p, q, a)$ tuples.

\paragraph{Off-the-shelf LLMs.} We benchmark models from three categories: (i) general-purpose open-weight models from the Gemma (4B--31B) \cite{google2025gemma3} and Mistral (7B--24B) \cite{mistral2025small30} families; (ii) domain-adapted biomedical models (MedGemma-1.5-4B-IT \cite{google2026medgemma}, BioMistral-7B \cite{labrak2024biomistral}); and (iii) large proprietary models from the Gemini \cite{gemini25}, GPT \cite{openai2024gpt4omini}, Claude \cite{anthropic2024claude3}, DeepSeek \cite{deepseek2026v4}, and GLM \cite{glm5team2026glm5vibecodingagentic} families. Smaller models were self-hosted via vLLM \cite{kwon2023vllm}; others were queried through the OpenRouter API (see Fig.~\ref{fig:benchmark} for detailed model list). 
    
\paragraph{Supervised fine-tuned models.}
% We investigate two fine-tuning approaches: Low-Rank Adaptation (LoRA)~\cite{hu2022lora}, a parameter-efficient adaptation method, and full-weight training, which is more computationally expensive but potentially more expressive.
We investigate two fine-tuning approaches: Low-Rank Adaptation (LoRA)~\cite{hu2022lora} and full-parameter training.
Due to resource constraints, these methods are applied only to relatively small models.
Each training sample is formatted as a three-turn conversation: the  system prompt with patient history, the query as the user turn, and the ground-truth answer as the assistant turn.
Training minimizes cross-entropy on assistant tokens only; when the tokenized sequence exceeds 25K tokens, the oldest reports are progressively dropped to retain the most recent history.
All models are fine-tuned for max two epochs on an H100 GPU, with early stopping based on validation CE-loss and hyperparameters selected on the validation fold.

All models (off-the-shelf and fine-tuned) are evaluated on the held-out test folds of both datasets using a shared instruction-based prompt lightly optimized on the validation set via GEPA \cite{agrawal2026gepa}.
% Due to computational cost, \texttt{gemini-3.5-flash} is evaluated on a 4\% subset of the MIMIC ICU Notes test fold.

\subsection{Evaluation metrics}

To measure accuracy, we employ an \textbf{LLM-as-a-judge} (Gemini 2.5 Flash \cite{gemini25}) to compare predicted answers against reference answers obtained from future segments of patient trajectories.
The judge assesses semantic equivalence: a prediction is marked correct if it captures the key factual content of the reference, regardless of exact wording; refusals to answer by the model are marked incorrect.
We manually verify a subset of judgments, finding them to be sufficiently reliable.
We do not expect 100\% accuracy, as future-oriented queries are inherently stochastic and may admit multiple plausible outcomes.
We therefore interpret this metric as a way to \textit{compare} approaches rather than as an absolute measure of clinical correctness.

\subsection{Results}

\begin{figure}[t]
    \centering
    \includegraphics[width=0.78\linewidth]{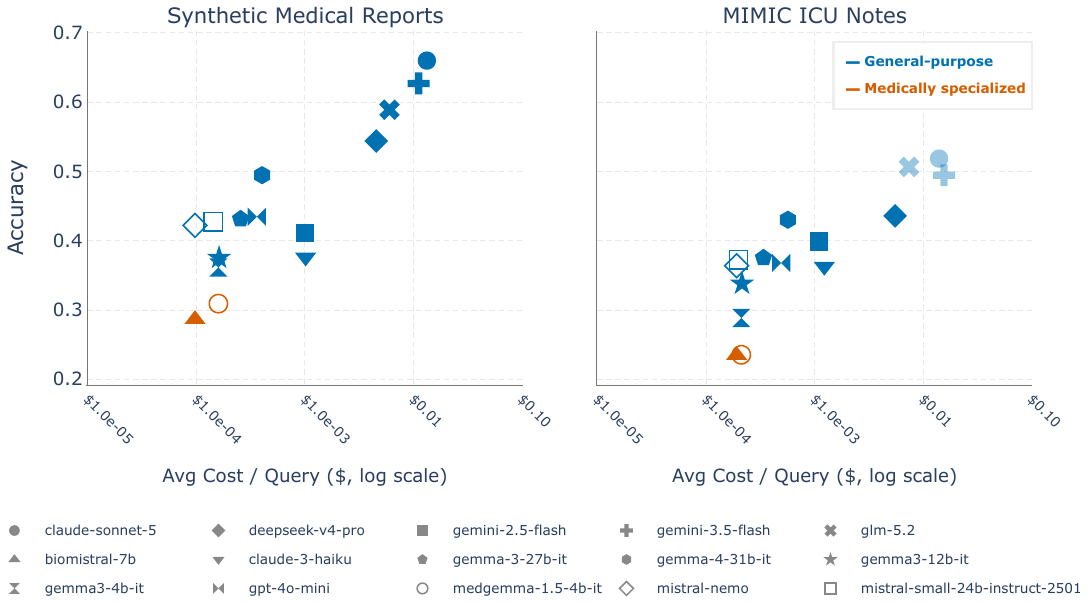}
   \caption{Baseline performance of off-the-shelf LLMs on the Synthetic Medical Reports and MIMIC ICU Notes benchmarks. The top legend row shows models with reasoning capabilities, evaluated using OpenRouter's default settings. For \texttt{gemini-2.5-flash} and \texttt{claude-sonnet-5}, these defaults did not enable reasoning. Due to computational costs, three models were evaluated on a 4\% subset of the MIMIC ICU Notes test set and are shown with reduced opacity.}
    \label{fig:benchmark}
\end{figure}

\paragraph{Performance of large-scale pre-trained models.}
To what extent does large-scale pre-training alone equip LLMs to answer future-oriented clinical queries? Fig.~\ref{fig:benchmark} visualizes the performance of various off-the-shelf LLMs with respect to their estimated cost, computed using average pricing from major LLM providers. We observe a clear trend: larger and more expensive models achieve higher accuracy on both datasets. Four frontier models consistently occupy the top positions on both benchmarks. On the Synthetic Medical Reports benchmark, all four surpass 50\% accuracy and are separated from the remaining baselines by a noticeable margin. On MIMIC ICU Notes, derived from real-world clinical records, only Claude Sonnet 5 and GLM-5.2 narrowly exceed this threshold. Notably, medically specialized models such as BioMistral-7B and MedGemma-1.5-4B-IT underperform compared to their general-purpose counterparts (Mistral-Nemo and Gemma-4B-IT), as well as most other evaluated models. We hypothesize that this reflects a mismatch between the medical task they have been fine-tuned on and the future querying task considered here.

\paragraph{Effect of fine-tuning.}

\begin{figure}[t]
    \centering
    \includegraphics[width=0.9\linewidth]{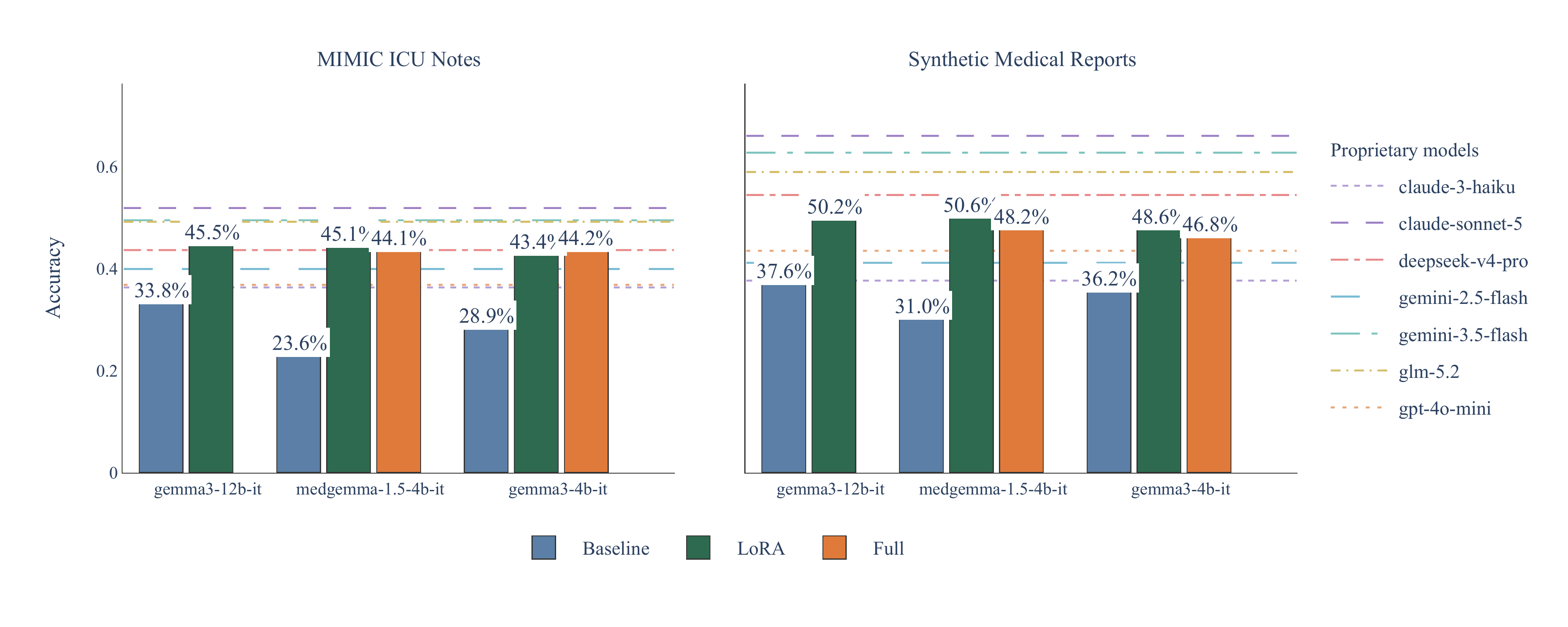}
    \caption{Effect of fine-tuning on future querying accuracy with respect to the accuracy of proprietary models.}
    \label{fig:training-results}
\end{figure}

Fig.~\ref{fig:training-results} compares three open-weight models before and after supervised fine-tuning on the future querying task. Training times ranged from 67.9 to 210\,h/epoch for MIMIC and 8.3 to 49.4\,h/epoch for Synthetic. LoRA (rank=16) adaptation yields consistent and substantial gains across all models and both datasets, ranging from +11.7 to +21.5 percentage points (pp) on MIMIC and +12.4 to +19.6\,pp on Synthetic. Notably, MedGemma-1.5-4B-IT, despite being the weakest baseline model, benefits the most from fine-tuning (+21.5\,pp on MIMIC, +19.6\,pp on Synthetic), reaching accuracy levels on par with the larger Gemma3-12B-IT. This suggests that task-specific supervision can largely compensate for the initial performance gap due to model size or prior specialization. After fine-tuning, all three models converge to comparable accuracy ranges (43--46\% on MIMIC, 46--51\% on Synthetic), indicating that the training signal, rather than model capacity, becomes the dominant factor for this task. Full-weight training, evaluated on MIMIC, provides a modest additional improvement over LoRA for Gemma3-4B-IT (+1.9\,pp) and small reduction for Medgemma-1.5-4b-it (-1.0\,pp). Full-weight training, evaluated on Synthetic data, shows decreased performance for both models compared to lora fine-tuning. This suggests that parameter-efficient adaptation captures most of the achievable gains. These results are encouraging for privacy-preserving deployments: relatively small, locally trainable models can approach the performance of larger proprietary systems evaluated in the previous section. However, the latest models still reach higher performances.

\vspace{-9pt}

% \begin{figure*}[t]
% \centering
% \begin{subfigure}[t]{0.4\textwidth}
%     \centering
%     \includegraphics[width=\textwidth]{figures/qa_prompt_figure.png}
%     \caption{FQ Inference Prompt}
%     \label{fig:qa-prompt}
% \end{subfigure}
% \hfill
% \begin{subfigure}[t]{0.4\textwidth}
%     \centering
%     \includegraphics[width=\textwidth]{figures/judge_prompt_figure.png}
%     \caption{LLM-as-a-Judge Prompt}
%     \label{fig:judge-prompt}
% \end{subfigure}
% \caption{Prompt structures used in our evaluation pipeline. (a)~The FQ inference prompt embeds the patient's chronological medical reports in the system prompt and poses the clinical query as the user message. (b)~The LLM-as-a-judge prompt compares the predicted answer against the ground truth and returns a binary correctness verdict.}
% \label{fig:prompts}
% \end{figure*}

\section{Discussion and conclusion}

Our results suggest that simple fine-tuning strategies can yield strong performance on the future-querying task, with small open-weight models matching larger proprietary systems.
This is an encouraging result for the development of privacy-preserving, on-premise deployments.
Direct comparison with prior benchmarks is not straightforward, as most existing approaches focus on predicting predefined outcomes at fixed time horizons from structured inputs.
In contrast, our framework addresses a broader setting, requiring models to answer open-ended queries over unstructured clinical text.
In this setting, multiple outcomes may be plausible and exact matches with a single ground-truth answer are not always expected, making traditional evaluation protocols and benchmarks only partially applicable.

A core strength of our approach is its task-agnostic nature: a single model operates directly on unstructured clinical documentation, requiring no manual feature engineering or structured data extraction pipelines, and can flexibly answer diverse clinical queries without retraining.
Key next steps include clinical validation with domain experts, establishing a human clinician upper bound, and distinguishing genuine world modeling from pattern matching (e.g., via consistency checks or counterfactual probing).
Beyond this, multimodal integration, uncertainty estimation, probabilistic judgments and robustness analysis remain important directions toward safe real-world adoption.

\vspace{-2pt}
\begin{credits}
\subsubsection{\ackname} The authors would like to thank Peter Peumans, Jean-Emmanuel Bibault, and Thomas Nedelec for their valuable insights and discussions.

\subsubsection{\discintname}
The authors have no competing interests.
\end{credits}

\newpage
\bibliographystyle{splncs04}
\bibliography{main}

\end{document}